\documentclass[fleqn,10pt]{article}

\usepackage[T1]{fontenc}
\usepackage{geometry}
\usepackage{authblk}
\usepackage{braket}
\usepackage{comment}
\usepackage{float}
\usepackage{graphicx}
\usepackage{xcolor}
\usepackage{amsmath}
\usepackage{amssymb}
\usepackage{setspace}
\usepackage{multirow}
\usepackage[colorlinks=true, linkcolor=blue, urlcolor=blue, citecolor=blue]{hyperref}

\title{\textbf{Efficient Semi-Automated Material Microstructure Analysis Using Deep Learning: A Case Study in Additive Manufacturing}}
\date{}

\author[1]{Sanjeev S. Navaratna}
\author[1]{Nikhil Thawari}
\author[1]{Gunashekhar Mari}
\author[1]{Amritha V P}
\author[1]{Murugaiyan Amirthalingam}
\author[1,2,*]{Rohit Batra}

\affil[1]{Department of Metallurgical and Materials Engineering, Indian Institute of Technology Madras, Chennai 600036, India}
\affil[2]{Center for Atomistic Modelling and Materials Design, IIT Madras, Chennai 600036, India}
\affil[*]{Corresponding author: rbatra@smail.iitm.ac.in}

\begin{document}

\maketitle

\begin{abstract}
Image segmentation is fundamental to microstructural analysis for defect identification and structure–property correlation, yet remains challenging due to pronounced heterogeneity in materials images arising from varied processing and testing conditions. Conventional image processing techniques often fail to robustly capture such complex features, rendering them ineffective for large-scale analysis. Even deep learning approaches struggle to generalize across such heterogeneous datasets owing to scarcity of high-quality labeled data. Consequently, segmentation workflows rely heavily on manual expert-driven annotations, which is labor-intensive and difficult to scale. Here, using the example additive manufacturing (AM) dataset, we present a semi-automated active learning-based segmentation pipeline that integrates a U-Net based convolutional neural network with an interactive user annotation/correction interface and a representative core-set image selection strategy. While the active learning workflow iteratively updates the model by incorporating user-corrected segmentations into the training pool, the core-set selection strategy identifies representative images to be considered in each iteration. Three subset selection strategies, i.e., manual, uncertainty-driven, and our proposed maximin Latin hypercube sampling from embeddings (SMILE) approach were evaluated over six refinement rounds with the SMILE method consistently outperforming other approaches and improving the macro F1 score from 0.74 to 0.93 on segmentation task, while reducing manual annotation time by approximately 65\%. The segmented defect regions were further analyzed using a coupled classification model, which categorized defects based on their microstructural characteristics. These classified defects were then mapped to corresponding AM process parameters, enabling insights between processing and defect types. Overall, the proposed framework substantially reduces labeling effort while maintaining scalability and robustness, and is broadly applicable to image-based analysis across diverse materials systems.
\end{abstract}


\section{Introduction}

An important goal in materials science is to understand how processing conditions influence microstructure and ultimately determine material performance \cite{olson1997computational,kumar2021methods,vora2020comprehensive,srivastava2022additive}. Important material properties such as strength and fatigue life are strongly affected by underlying microstructural defects \cite{suresh1998fatigue}. Porosity and lack of fusion defects often act as stress concentrators and control how failure occurs in metals, ceramics, and composites \cite{anderson2005fracture}. Hence accurate defect detection and classification is essential for establishing reliable structure-property relationships \cite{withers2021x}. Microstructural characterization is commonly performed using optical and scanning electron microscopy, followed by image segmentation to quantify defect types and features (such as density, morphology, etc.)
In practice, segmentation is typically carried out using classical image processing techniques such as global thresholding or Otsu’s method \cite{sezgin2004survey}. While effective for small and simple datasets, these approaches lack robustness under varying imaging conditions and do not scale to large, heterogeneous image databases \cite{decost2015computer}.

These limitations are especially pronounced in additively manufactured (AM) materials. While additive manufacturing enables complex geometries through layer-by-layer fabrication \cite{debroy2018additive,parimi2014microstructural,francois2017modeling,song2018numerical}, the localized heat input and rapid solidification inherent to the process introduce defects such as porosity and lack of fusion\cite{king2014observation}.These defects strongly influence the mechanical properties and fatigue behavior of AM components \cite{grasso2017process}, making reliable defect quantification essential for establishing meaningful process–structure-property relationships \cite{grasso2017process}. Furthermore, AM microstructures for a given material composition exhibit substantial variability in defect size, shape, and contrast, driven by changes in process parameters such as laser power and scan speed \cite{zerbst2021damage,gordon2020defect}, thereby making it difficult to develop a universal and robust defect detection strategy.

Several prior studies have investigated defect quantification in AM materials using image based analysis and supervised learning techniques. For example, Ellendt et al.~\cite{ellendt2021poreanalyzer} developed a rule-based automated framework for pore detection that uses predefined geometric features such as area, perimeter, aspect ratio, and circularity of defects to classify and correlate them with process parameters. While this approach performs well for small datasets, its dependence on predefined geometric features limits its ability to classify complex or highly irregular defect morphologies. Altmann et al. \cite{altmann2023defect} extended this approach by combining additional geometric descriptors, such as aspect ratio, circularity and solidity, with supervised machine learning (ML) models (e.g., random forest).  

Although this resulted in improved performance, the approach still relied on hand-crafted features and well curated training data. More recently, Wong et al. \cite{wong2022segmentation} introduced a U-Net–based deep learning framework for automatic segmentation of internal defects in AM components using X-ray computed tomography data. By formulating defect detection as a 2D pixel level and 3D voxel level segmentation task, their work demonstrated that CNNs can learn complex pore geometries directly from high resolution images. However, this framework was developed and validated on a single material system with a limited number of specimens and relied on labels generated using a thresholding scheme, raising concerns about model generalizability and label quality.

Overall, the existing studies on defect detection or classification implicitly assume availability of sufficiently large, high quality annotated datasets \cite{ma2023review}, which severely constrains scalability and broader adoption of these models. On a related note, models trained on a single dataset often fail to generalize across different materials, microstructures, or imaging conditions, leading to degraded performance outside the original training domain for which annotated datasets are available \cite{li2025novel}. Another important aspect, especially in the context of AM materials, is that these approaches do not explicitly incorporate microstructural context into defect classification. Etched microstructural images provide essential complementary information such as melt pool boundaries, grain morphology, and phase contrast that is critical for accurate interpretation of defect types. For example, lack of fusion defects are commonly associated with irregular melt pool overlap near grain boundaries and partially unmelted regions, whereas porosity typically appears more circular and isolated. Thus, accurate defect classification requires models that leverage multimodal image data from both polished and etched samples.

Thus, to enable efficient and scalable data-driven defect detection and classification we make following two key observations. First, annotating a diverse and representative subset of images is sufficient to capture variability in defect morphology, as many microstructural images contain redundant or overlapping defect patterns. Therefore, selecting images that span range of defect sizes, shapes, and textures can allow the model to learn the underlying defect distribution while minimizing redundant labeling effort \cite{qiu2025decomposition, kaushal2018learning, durga2021training}.

Second, active learning approaches coupled with a human-in-the-loop framework can substantially reduce manual annotation effort by iteratively selecting the most informative samples for annotation and incorporating expert feedback into the subsequent training cycles \cite{amin2023deep,kale2025active}. Such approaches are particularly well suited for materials science applications, where potential candidate space is large but labeling cost is high \cite{mittal2025realistic}, and have previously been explored through uncertainty-based sampling strategies for segmentation tasks \cite{gaillochet2023active}. 
Based on these observations, here we present a semi-automated, two-stage pipeline for defect quantification consisting of defect detection followed by defect classification. The defect detection stage integrates an active learning framework with a U-Net based CNN, an interactive annotation interface, and a core-set selection strategy. The key idea is to move away from treating segmentation as a single-shot task requiring large labeled datasets, and instead progressively refine the model through iterative training cycles on small, manageable subsets of data. Further, in each active learning cycle, the user need not perform full annotations to generate ground truth labels. Instead preliminary labels are obtained from the current model, which are corrected by the user using an easy-to-use annotation interface, thereby quickly generating the next round of training data. This reduces the user’s task to error correction rather than full-scale annotation, substantially lowering labeling time and cost. Furthermore, as training advances, prediction accuracy improves and the amount of manual correction required decreases. In regard to the images to be considered in each active learning loop, besides baseline manual and uncertainty based sampling strategies, we introduce a new subset selection method termed sampling using maximin–Latin hypercube sampling from embeddings (SMILE). This method leverages feature embeddings to explicitly promote coverage and diversity among selected image samples, ensuring that user effort is focused on a representative subset of the dataset. Post segmentation, detected defect regions in newly acquired AM material images are extracted as image patches, each corresponding to a defect instance. The size of the image patch is generated in such manner that along with the defect neighboring microstructural context is also captured. These patches are then given as input to a custom CNN model for defect classification. Finally, the classified defects are linked to their corresponding AM process parameters, such as laser power and scan speed, enabling quantitative analysis of the relationships between processing conditions and defect formation. Although demonstrated here for AM materials, the proposed modular pipeline is general and provides an efficient, scalable and data-driven framework for defect detection and classification across a wide range of materials manufacturing processes.

\section{Methodology}

In this work we address the problem of AM defect quantification as a sequential two-stage problem, i.e., defect detection followed by defect classification. While the first problem requires only polished sample images, the latter additionally requires etched images for microstructural context. Below we provide more details on the different aspects of the proposed methodology.

\begin{figure}[t]
    \centering
    \includegraphics[width=1\textwidth]{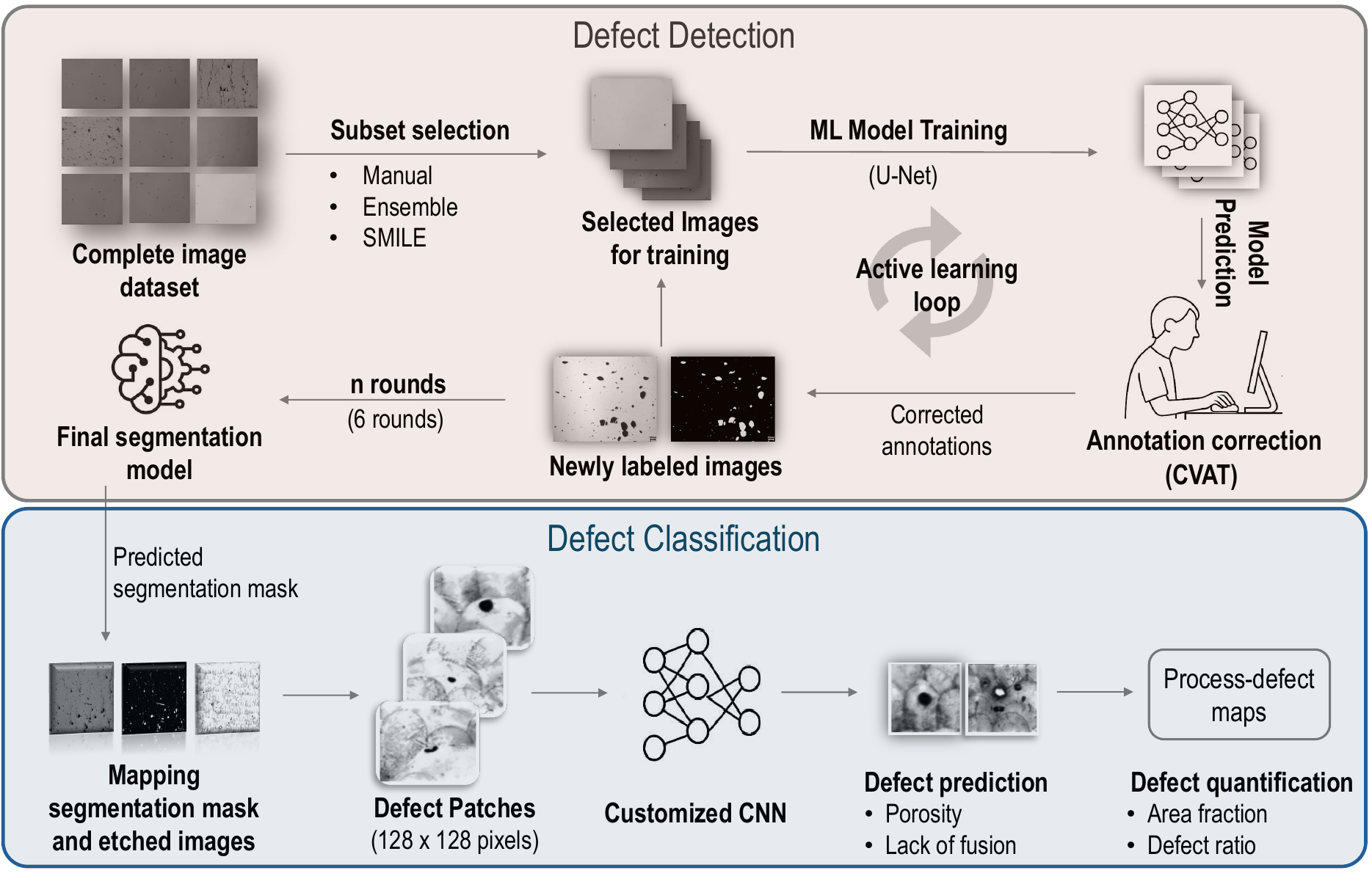}
    \caption{\textbf{Semi-automated pipeline for microstructure image analysis.} The workflow consists of two stages: defect detection and defect classification. In the defect detection stage, an active learning framework combined with a subset selection strategy and a user-friendly annotation interface is employed to segment regions of microstructural images containing defects. The subset selection strategy ensures the use of diverse and representative images, active learning enables iterative improvement of the detection model, and the annotation tool allows users to efficiently correct model-predicted labels rather than generating ground-truth annotations from scratch. Subsequently, in the defect classification stage, the obtained segmented defect regions are mapped to corresponding etched microstructural images, containing critical microstructural features such as melt pool boundaries and grain morphology. This combined information is used to train a defect classification model, and the resulting defect classification information is mapped to associated manufacturing process parameters.}
    \label{fig:methodology}
\end{figure}

\subsection{Dataset}

The imaging dataset was generated by capturing optical micrographs of cube samples (10 × 10 × 10 mm³) fabricated using the Laser Powder Bed Fusion (LPBF) process with different alloy grades \cite{wang2018scanning,gong2014analysis}. For sample preparation, the cubes were first sectioned diagonally and then hot mounted. The mounted specimens were polished to achieve a mirror-like surface finish using emery papers ranging from 320 to 2500 grit, followed by final polishing on a velvet cloth with an alumina slurry. The polished samples were examined under an optical microscope to identify internal defects such as porosity and lack of fusion. Images were acquired at different magnifications (e.g., 5×, 10×, and 50×) and under varying imaging conditions, resulting in differences in contrast, field of view, and spatial resolution ($\mu$m/pixel) \cite{laleh2021critical}. Additionally, the acquired images had varying pixel dimensions due to variation in acquisition settings.

For defect detection, 67 polished optical images, captured from Inconel 625, Al7075 aluminum alloy, and HK30 steel samples, were used for model training. An additional 30 polished images were kept reserved as an independent test dataset for evaluating model effectiveness \cite{pleass2018influence,mao2024research}. For defect classification, 71 pairs of polished and etched images for CoCrMo alloy, and 16 pairs of images for Inconel 625 were collected using the same imaging modality \cite{aoyagi2019simple}. The etched sample images were prepared under varying process parameters to reveal microstructural features such as melt pool boundaries and grain structures. The etchants were chosen based on literature and optimized experimentally for each specific alloy \cite{sommer2022revealing,shafiee2023alternative}. The classification dataset was further mapped to the corresponding AM process parameters to enable the analysis of processing–defect relationships {\cite{dogu2022digitisation}}.

\subsection{Defect Detection}
In the first stage, defect detection was carried out using a semi-automated segmentation pipeline as shown in Figure 1 (top panel). The objective was to accurately localize defects in microstructural images. Hence, the task was formulated as a binary semantic segmentation problem, with all defects treated as a single foreground class and the remaining regions as background, consistent with common practices in segmentation studies \cite{stuckner2022microstructure,ronneberger2015u}.

As stated earlier, this stage focuses on two key aspects: the selection of representative and diverse images for training using a core-set selection strategy, and the use of an active learning workflow to reduce manual annotation effort. The central idea is to iteratively improve the segmentation model while minimizing the manual labeling effort. Accordingly, the detection pipeline followed an active learning procedure in which a small, informative subset of images was first manually annotated to generate initial ground truth masks and train the model. The trained model was then applied to unlabeled images to produce preliminary defect masks, which were reviewed and corrected by a human expert using an open-source computer vision annotation tool (CVAT) \cite{team2023cvat}. These corrected samples were added back to the training set, and the model was retrained on the expanded dataset. This cycle was repeated until segmentation performance was found to converge.

A key aspect of active learning is the selection of images used for training in each iteration. Numerous subset selection strategies have been proposed to prioritize informative and diverse samples while minimizing redundancy. It has been shown that carefully selected subsets can achieve comparable performance with significantly fewer labeled examples \cite{siddiqui2020viewal}. In this work, we evaluated two commonly used selection strategies, namely, manual selection and deep ensembles, as baseline against our proposed core-set selection algorithm, termed SMILE.
Manual selection is the simplest and most intuitive strategy that has been widely used in prior active learning studies \cite{settles.tr09}, where samples are chosen based on visual inspection and prior experience. In this work, images exhibiting diverse defect sizes, shapes, and contrasts were manually grouped, and subsets from these groups were selected for training. While this approach leverages domain knowledge and can be effective when defects are rare, it does not scale well to large datasets and is inherently subjective, potentially introducing user bias. Thus, as another baseline, uncertainty-based sampling was also adopted, wherein images with the highest model prediction uncertainty were selected for subsequent active learning iterations. For this, uncertainty was first computed at the pixel level using predictions from an ensemble of 10 U-Net models and was then averaged to compute the image-level uncertainty, a strategy commonly applied in semantic segmentation \cite{lakshminarayanan2017simple, beluch2018power}. Henceforth, this strategy is termed as the ensemble approach.

Finally, as discussed above, we introduce SMILE, a novel core-set based subset selection strategy. Prior studies have shown that selecting a diverse core-set in feature space can significantly improve batch active learning for deep neural networks by minimizing the maximum distance between unlabeled samples and the selected subset, thereby promoting broad coverage of the feature space and outperforming uncertainty-based methods in batch settings \cite{sener2017active, ash2019deep}. Building on this idea, in the SMILE algorithm the complete set of unlabeled images was first projected into a two-dimensional embedding space using t-SNE. This dimensionality reduction preserved local neighborhood relationships, meaning samples similar (or dissimilar) in the original feature space remained closer (or farther) to each other in the embedding space. K-means clustering was then applied to the image embeddings to partition the data into groups corresponding to distinct regions of the feature space, thereby grouping images with redundant information. The number of clusters was determined using the silhouette score, which measures cluster cohesion and separation \cite{rousseeuw1987silhouettes}.  Next, for each cluster, the standard deviation along both embedding dimensions was computed to quantify intra-cluster dispersion. For this, a single variability metric, spread, was computed which is defined as the euclidean norm of the standard deviation vector \cite{johnson2007applied}. Clusters were then ranked in the descending order of this spread, and samples within each cluster were selected using LHS with maximin criterion to be included in the training dataset in each active learning iteration. Together, these steps enabled efficient selection of diverse and representative samples from the complete dataset.

For each subset selection strategy, six active learning rounds were performed. In each round, four images were selected using the chosen strategy and were added to the training set, after which the segmentation model was retrained on the expanded dataset. A fixed set of 30 images (unetched polished) was reserved as an independent test set for all models to enable consistent comparison across different subset selection strategies.
Within the active learning framework, defect segmentation was performed using a U-Net architecture implemented in the AtomAI framework \cite{vashishtha2025reusability}. The encoder–decoder structure with skip connections enabled the network to capture both fine defect details and broader microstructural context, which is critical for detecting low-contrast and irregular defects \cite{ronneberger2015u}. U-Net–based models have been widely applied to materials imaging, and prior work has demonstrated their effectiveness using the patch-based training scheme \cite{vashishtha2025reusability}. Motivated by these results, the same architecture and similar training strategy was adopted here. In each active learning round, one image was randomly set aside as a test image, while the remaining were used to create smaller patches of size 256$\times$256 pixels. For every training image, 500 patches were generated using AtomAI. These patches were then divided into an 80-20 train-validation split. Data augmentation was applied during training patch construction and included random rotations and spatial transformations. Binary cross-entropy was used as the loss function, and early stopping was employed to prevent overfitting and ensure stable convergence. Model performance was evaluated using precision, recall, and macro F1 score on the fixed independent set of 30 images. The macro F1 score assigns equal importance to defect and background classes, which is critical for reducing manual annotation effort, as poor performance on either class increases the need for human correction. More details on hyperparameters used to train the defect segmentation model are provided in the Supporting Information (SI), along with the obtained training and validation loss curves in each active learning iteration.

To facilitate model refinement, annotation correction was performed using CVAT \cite{team2023cvat}, which enables efficient visualization and editing of predicted annotations. The trained segmentation models were deployed using the serverless platform Nuclio \cite{nuclio}, allowing CVAT to directly invoke the trained models and generate predictions within the annotation interface. The entire inference pipeline was containerized using Docker \cite{docker}. In each active learning round, all steps, including generation of model predicted labels for unlabeled images, expert review with addition of missing defects and removal of false detections, and saving of the final ground truth masks, were performed within CVAT.

\subsection{Defect Classification}

The second stage (Figure 1, bottom panel) focused on defect classification by combining the segmented defect masks from the first stage with additional microstructural information obtained after chemical etching. To achieve this, each unetched image was paired with its corresponding etched image acquired from the same field of view, ensuring that defect locations remained approximately aligned between the two modalities. This enabled direct transfer of the segmentation masks generated using the best performing model from the first stage to the etched images. Once the defect masks were obtained, individual defect instances were then extracted following a prior approach \cite{vashishtha2025reusability} that uses contour-based connected region extraction, as implemented in OpenCV \cite{opencv_library}. With each defect instance identified, a fixed-sized 128$\times$128 patch was generated around the defect center. While this strategy was directly applicable to smaller defects ($<$128 pixels) without loss of local microstructural context, for larger defects, bounding boxes padded by 10 pixels were first used to fully enclose the defect region, and the resulting patches were subsequently resized to 128$\times$128 pixels. This strategy accommodated defects of varying sizes while ensuring consistent input dimensions for subsequent CNN-based classification.

In total, 233 defect-centered patches were extracted from 16 etched images of Inconel 625 samples fabricated under varying processing conditions. Of these, 23 patches were reserved as an independent test set, while the remaining samples were used for training. A customized CNN architecture was employed to classify defects as either porosity or lack of fusion, and was initialized with ImageNet-pretrained weights to leverage transfer learning for improved feature representation \cite{deng2009imagenet}. To address class imbalance, focal loss was used during training \cite{lin2017focal}. The CNN classification model trained on Inconel 625 dataset was directly used for predictions on CoCrMo alloy dataset (71 polished-etched image pairs) without any modifications. More details on hyperparameters used to train the defect classification model are provided in SI, along with the obtained training and validation loss curves.

In the final stage of the workflow, the classified defects were mapped to their corresponding process parameters. Laser power and scan speed were obtained from the build data and linked to each image. Using the classification model, defect statistics were computed for each image, including the defect area fraction (total defect area normalized by image area), total defect count, and the fraction of each defect type for a given process condition. These metrics were averaged across all images acquired under the same parameter combination to enable consistent comparison between conditions. Such statistical measures are commonly used to relate defect area fraction and morphology to processing variables in AM materials \cite{sanaei2019defect}. The resulting process-defect maps provide a systematic and quantitative view of how laser parameters influence defect formation across both material systems, thus enabling optimization of processing parameters.

\section{Results and Discussion}

\subsection{Defect Detection Performance}

Figure 2a illustrates the improvement in segmentation performance across successive active learning rounds, measured by the macro F1 score on the independent test set; error bars denote $\pm1\sigma$ standard deviation in the macro F1 score computed over the 30 test images. The initial U-Net model can be seen to have only moderate performance as it is trained on a limited annotated dataset. Further, defect segmentation in AM samples is known to be challenging because defects often exhibit low contrast, irregular shapes, and subtle boundaries that blend with the background \cite{ronneberger2015u}. This underscores the importance of increasing both the volume and diversity of training data. All subset selection strategies show consistent performance gains with each round, demonstrating the effectiveness of the iterative active learning framework. These improvements arise from the progressive increase in both the size and diversity of the labeled dataset.

As a baseline, classical Otsu thresholding was also evaluated. Otsu’s method relies on global intensity thresholding to separate foreground and background regions. While computationally inexpensive and training-free, it performed poorly for defect detection, achieving a macro F1 score of 0.83 compared to 0.94 for SMILE in the final round. Consequently, the masks generated by Otsu required substantial manual correction, negating any potential gains in annotation efficiency.

\begin{figure}[!htb]
    \centering
    \includegraphics[width=1\textwidth]{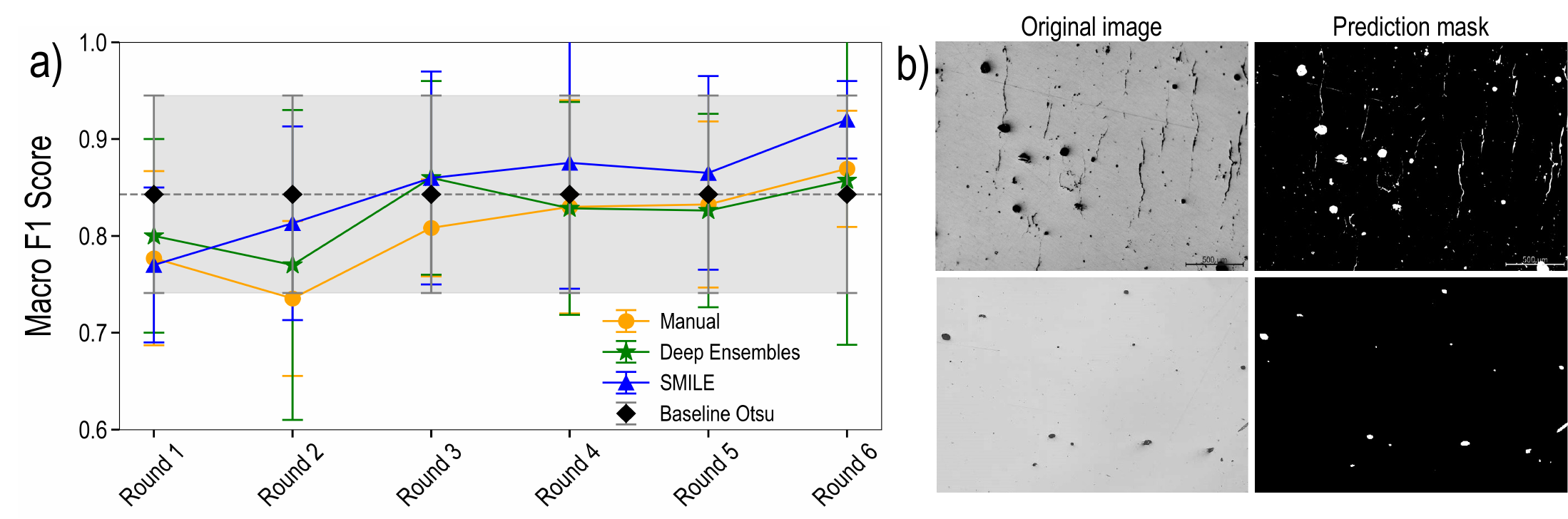}
    \caption{\textbf{Defect detection model performance.}
    (a) Iterative improvement in the macro F1 score of the defect detection model with increasing active learning rounds for different subset selection strategies. Results obtained using the baseline Otsu thresholding method are also shown for comparison. Markers denote the mean performance on the test set, and error bars indicate $\pm1\sigma$ standard deviation. The dashed line and shaded region represent the mean macro F1 score and $\pm1\sigma$ standard deviation of the Otsu method, respectively. (b) Example segmentation results produced by the final best-performing SMILES-based model on test images.
}
    \label{fig}
\end{figure}

Comparative performance of different subset selection strategies, i.e., manual sampling, deep ensemble and SMILE, used during active learning can also be seen in Figure 2a. Manual sampling leads to slower and less stable improvements in segmentation performance. This behavior is expected because, although manual selection may account for factors such as defect type, defect area fraction, and morphology, it remains primarily visual rather than quantitative. It lacks rigor and may result in inclusion of redundant samples while overlooking images that contain unique or less frequent characteristics. Similar limitations of manual sampling have been reported in earlier active learning studies \cite{mittal2025realistic}. Ensemble-based sampling improves upon this by prioritizing samples with high predictive uncertainty, which focuses annotation effort on difficult regions and produces more consistent gains. However, the improvement tends to saturate in later refinement rounds, likely because uncertainty becomes concentrated around a limited set of ambiguous patterns, resulting in redundant selections and incomplete coverage of the overall data distribution.

Among all evaluated strategies, the SMILE-based approach achieved the highest macro F1 scores consistently across all active learning rounds. Unlike uncertainty-based methods, SMILE explicitly promotes coverage and diversity in the feature embedding space using K-means clustering, and Latin hypercube sampling combined with a maximin criterion. This encourages selection of samples that are well distributed across the dataset, allowing the model to observe a wider range of defect sizes, shapes, and background microstructures during training.  This behavior is illustrated using the two-dimensional Isomap visualization in Figure 3a, which shows the embeddings of the full training set (black circles) and the independent test set (purple crosses), along with samples selected based on different approaches overlaid using different marker styles. First thing to be observed is the uneven coverage of the data in the embedding space, characterized by several dense clusters alongside more diffuse, sparsely populated, and isolated regions. This highlights the inherent challenge in developing a generalizable model for heterogeneous materials datasets. More importantly, distinct patterns of underrepresented and overrepresented regions across different subset selection strategies can also be noticed (indicated by red rectangles). For instance, the top-left region is clearly underrepresented by manual selection despite containing several test points. Similarly, the region immediately below can be seen to have relatively sparse coverage by the ensemble-based approach. In contrast, the central left region in the plot shows clear oversampling by both the manual and ensemble strategies, with many selected samples concentrated within an already dense cluster. On the contrary, the SMILE-based approach demonstrates a more balanced distribution of selected samples across both dense and sparse regions of the embedding space. This uniform coverage further supports the effectiveness of the SMILE-based strategy, while highlighting the representational limitations of the manual and ensemble-based selection methods.

\begin{figure}[p]
    \centering
    \includegraphics[width=\textwidth,height=0.75\textheight,keepaspectratio]{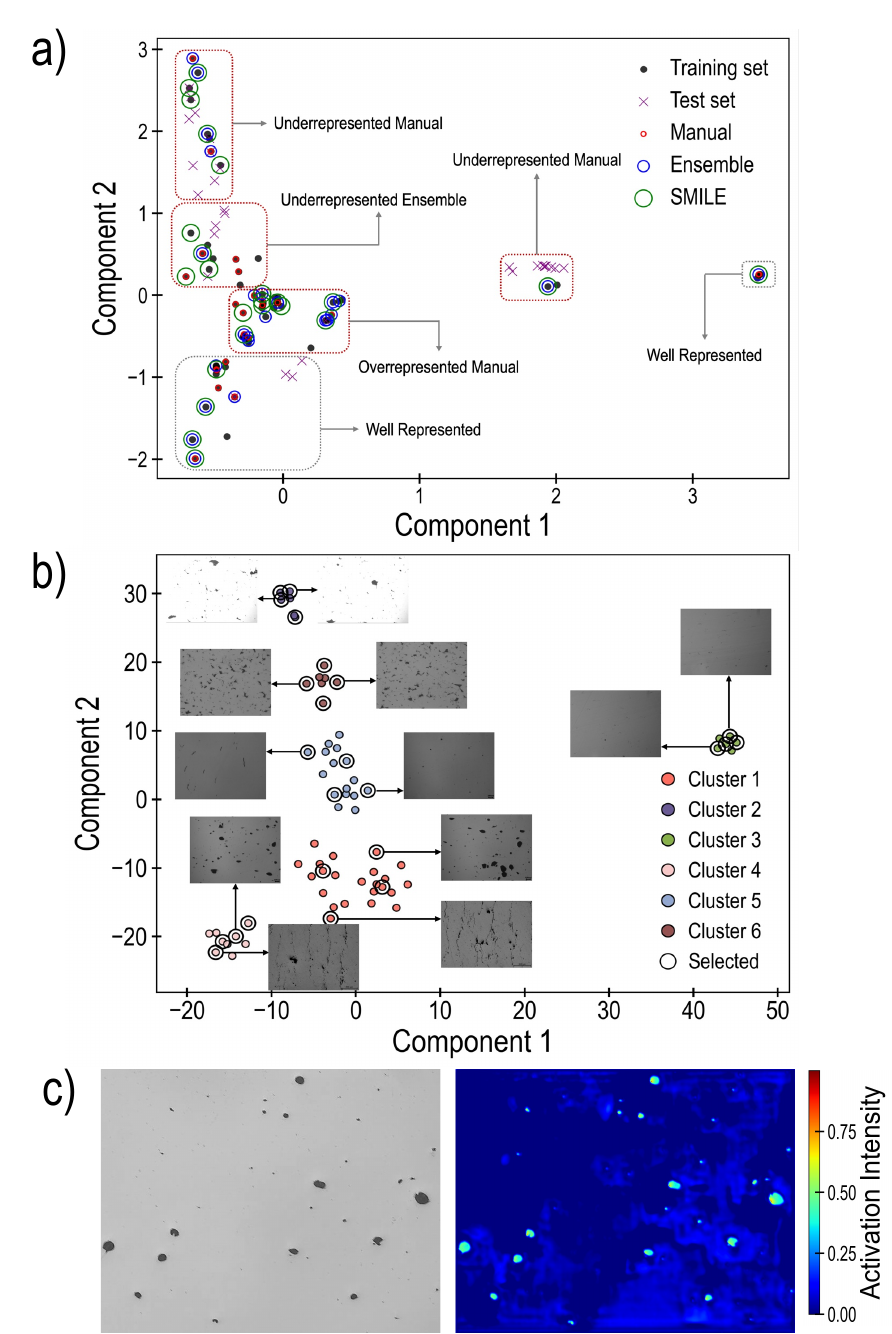}
    
    \caption{\textbf{Diversity and coverage of subset selection strategies and Representative Grad-CAM visualization using the final SMILE-trained model.} (a) Isomap projection of the full training and the independent test set, highlighting image samples selected using manual, deep ensemble, and SMILE-based approaches across all active learning iterations. Regions that are either underrepresented or overrepresented by different selection strategies are also highlighted. (b) t-SNE projection of the image samples selected using the SMILE approach. While selection of points across various clusters promotes coverage of the embedding space, the LHS + maximin criteria enforces intra-cluster diversity. (c) The left panel shows the original microstructural image, while the right panel displays the Grad-CAM heatmap overlaid on the image. High activation regions are localized around defect areas, confirming that the network focuses on physically meaningful microstructural features rather than background intensity variations. }
    
    \label{fig1}
\end{figure}

The t-SNE visualization in Figure 3b further highlights efficient coverage achieved by SMILE across diverse defect morphologies. The SMILE selected samples can be seen to be distributed across different regions of the embedding space. This behavior reflects the design of the SMILE algorithm, in which clustering enforces global coverage, while LHS with maximin criterion promotes intra-cluster diversity. This effect was further quantified using the Wasserstein distance \cite{rubner2000earth} between the embedding distribution of the subset selected 24 images (across six active learning rounds) and that of the full dataset. While manual sampling showed the largest distance mismatch (0.96), uncertainty-based sampling showed significant improvement (0.84), but SMILE achieved the lowest distance (0.82), indicating the closest match to the full data distribution. This reduced distributional gap explains the superior segmentation performance of SMILE and highlights the benefit of diversity-aware sampling for more robust, generalizable and efficient learning \cite{sener2017active}.

Representative qualitative segmentation results from the final SMILE model for two images from the independent test set are included in Figure 2b. The model can be seen to accurately capture both small and large defects, delineate clear boundaries, and exhibit few false positives. To further assess whether the model learned physically meaningful features, gradient-weighted class activation mapping (Grad-CAM) was employed \cite{selvaraju2017grad}. Grad-CAM heatmaps, computed using the pre-final convolutional layer of the final SMILE model as observed in Fig. 3(c), were strongly localized around defect regions, while background areas consistently showed low activation. This clearly indicate that the network learned to focus on defect-related structures, including both interiors and boundaries, rather than relying on global intensity variations. These results confirm that the strong performance of the SMILE-based U-Net arises from meaningful and interpretable feature learning.

While the preceding sections focused on segmentation performance, the primary practical benefit of the proposed approach lies in the reduction of annotation effort. Annotation efficiency was evaluated by comparing the time taken for full manual labeling against model-assisted correction in CVAT using predictions from the final SMILE model. Three student annotators (A1–A3) independently labeled five representative images spanning varying defect densities and microstructural backgrounds. As summarized in Table 1, manual annotation required 140, 105, and 93 minutes for A1–A3, respectively, whereas model-assisted annotation reduced this to 56, 35, and 28 minutes, corresponding to an average time saving of approximately 65\%. These results demonstrate that combining active learning with SMILE-based subset selection substantially reduces expert effort while maintaining high segmentation quality. This reduction is particularly important for large microstructural datasets or images with high defect densities (e.g., dislocations, twins, or grain boundaries), where annotation time is a major bottleneck. Overall, the proposed semi-automated framework provides a scalable and efficient solution for defect detection in practical materials characterization workflows.

\begin{table}[!ht]
\centering
\small
\caption{\textbf{Annotation time savings}. Comparison of manual and model-assisted annotation time (in minutes)  across three annotators (A1–A3) for five test images. The model-assisted approach reduced the annotation time by an average of 65\%.}
\begin{tabular}{c|ccc|ccc}
\hline
\hline
\multirow{2}{*}{\textbf{Image No.}} &
      \multicolumn{3}{c|}{\textbf{Manual (min.)}} &
      \multicolumn{3}{c}{\textbf{Model (min.)}} \\
      \cline{2-7}
 & 
\textbf{A1} &
\textbf{A2} &
\textbf{A3} &
\textbf{A1} &
\textbf{A2} &
\textbf{A3} \\
\hline
1 & 45 & 35 & 30 & 12 & 7 & 7 \\
2 & 30 & 25 & 25 & 20 & 13 & 10 \\
3 & 25 & 15 & 15 & 15 & 9 & 5 \\
4 & 30 & 20 & 15 & 8  & 5 & 4 \\
5 & 10 & 10 & 8  & 1  & 1 & 2 \\
\hline
\textbf{Total} & 140 & 105 & 93 & 56 & 35 & 28 \\
\hline
\textbf{Mean} & \multicolumn{3}{c|}{112.67} & \multicolumn{3}{c}{39.67} \\
\hline
\hline
\end{tabular}

\end{table}

\subsection{Defect Classification Performance}

Next, the segmented defect regions obtained using the best SMILE-based model were further analyzed for defect classification. Unlike the segmentation stage, which treated all defects as a single foreground class, this step focused on classifying defects as either porosity or lack of fusion by incorporating additional microstructural information obtained after sample etching. The performance of the CNN-based classification model on the held-out test set is summarized by the confusion matrix in Figure 4a. The model achieved an overall accuracy of 0.87 and a macro F1 score of 0.86. In contrast, the same architecture trained without pretrained weights of ImageNet achieved a substantially lower macro F1 score of 0.68, highlighting the benefit of transfer learning. As an additional benchmark, the classification model was trained directly on the etched images omitting the segmentation stage. This model resulted in a macro F1 score of just 0.43, further demonstrating the advantage of the proposed two-stage framework.

Representative classified patches from the test set are shown in Figure 4b. The model correctly identified most defects despite variations in shape and surrounding microstructure, with only 3 misclassifications out of 23. Notably, the CNN classification model trained exclusively on the Inconel 625 dataset exhibited comparable performance on the CoCrMo alloy dataset and thus was directly used for subsequent analysis without any modification. This result also highlights the strong generalization capability of the proposed approach.
Overall, these results show that pairing accurate segmentation with microstructural context enables reliable defect classification and offers an efficient way to quantify defects for correlation with AM process parameters.

\begin{figure}[t]
    \centering
    \includegraphics[width=1\textwidth]{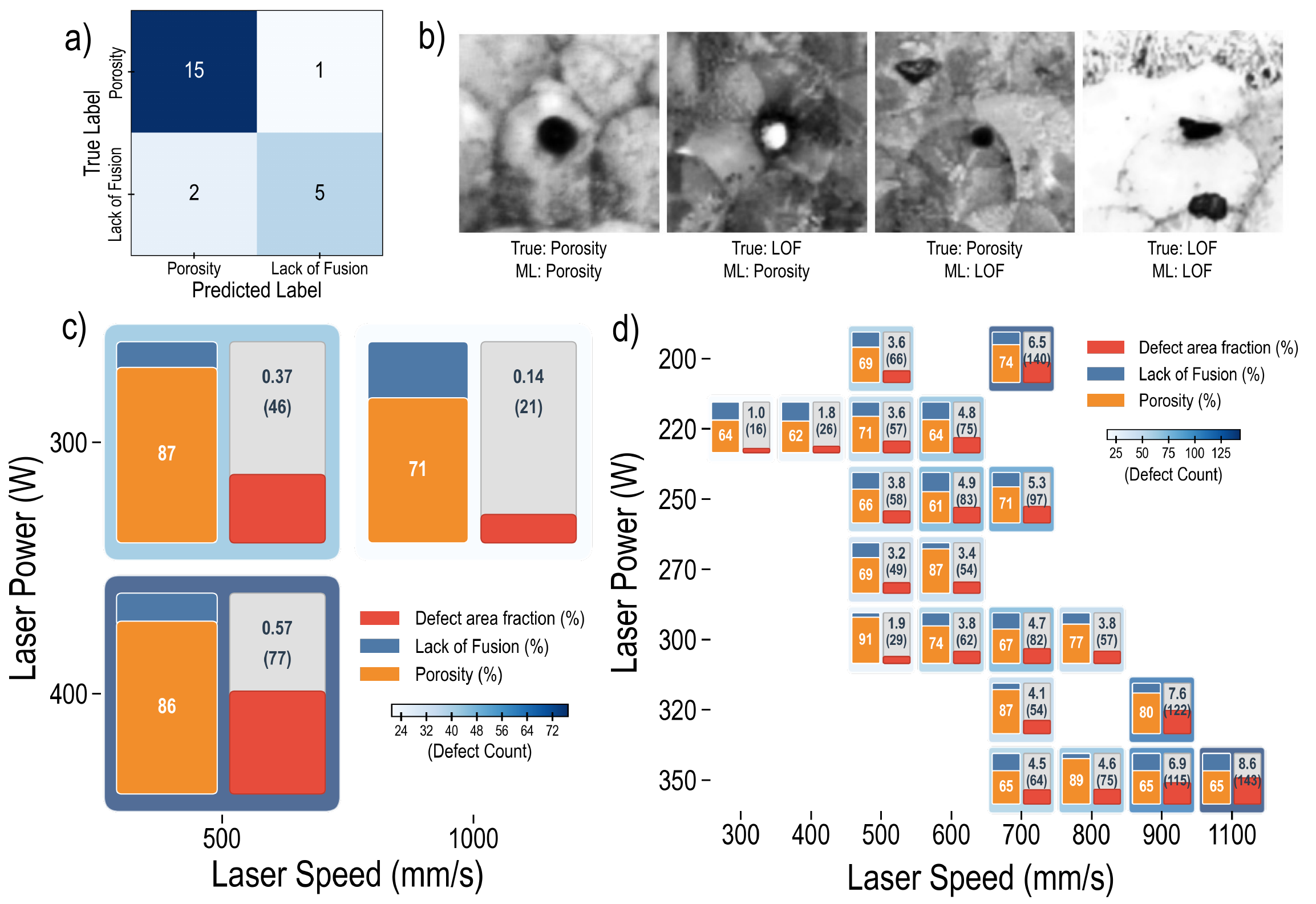}
    \caption{\textbf{Defect classification model performance and defect statistics measured across different AM process parameters.} (a) Confusion matrix summarizing performance of the defect classification model on the test set. (b) Representative test patches comparing ground-truth and predicted labels, illustrating both correct and incorrect classifications. Defect statistics measured across different laser power and scan speed combinations for (c) Inconel 625 and (d) CoCrMo alloy AM systems. Each cell represents one processing condition and contains two bar plots, one capturing relative fraction of lack of fusion (blue) and porosity (orange), with the white colored number indicating the porosity fraction, and second bar denoting \% defect area fraction (red color) with the average number of defects reported in brackets. The results highlight clear variations in defect type and severity across processing conditions, demonstrating the strong influence of process parameters on defect formation.
}

    \label{fig2}
\end{figure}

Following the classification described above, the detected defects in each image were aggregated to extract quantitative metrics, including defect count, defect area fraction, and the relative fractions of defect types, namely lack-of-fusion (LoF) and porosity. This analysis was performed for two AM material systems, Inconel 625 and CoCrMo alloy, with the resulting trends plotted as a function of laser power and scan speed in Figures 4c and 4d, respectively. Figure 4c presents processing conditions showing that porosity is the dominant defect type under higher energy input, while LoF defects are more prevalent at lower energy input conditions. Although, due to the limited coverage of the process parameter space in Fig. 4c, these observations alone do not fully capture the broader process–defect relationships. A more comprehensive trend is revealed in Figure 4d, where the upper region of the process map (laser power below 300 W) consistently shows a higher fraction of LoF defects regardless of laser speed, consistent with literature reports linking LoF defects to insufficient energy density and incomplete melt pool overlap \cite{shrestha2022formation}. In contrast, conditions above 300 W exhibit a more irregular distribution of LoF defects, likely resulting from variations in laser speed that modify the effective energy input and melt pool stability. Another interesting observation is porosity became more prominent at higher energy input conditions, where increased laser power and reduced scan speed both promote keyhole formation and less stable melt pools. These trends are consistent with previously reported studies on AM studies \cite{cacace2022effect}, further supporting the validity of the developed ML-based defect classification framework.

Further clear differences in defect area fraction and relative defect fractions are observed between the two materials in Figure 4. In particular, CoCrMo alloy exhibited higher defect sensitivity compared to Inconel 625, which can be attributed to material-specific thermophysical properties such as thermal conductivity and solidification behavior, as reported in prior studies \cite{mukherjee2016printability}. Beyond changes in defect type and area fraction, the total number of detected defects also varied markedly with processing parameters. For Inconel 625, the average defect count ranged from 21 to 77 across the investigated conditions, with the maximum observed at 400 W and 500 mm/s. In contrast, CoCrMo alloy showed a much wider variation, with defect counts ranging from 16 to 143, especially at higher laser power and scan speed combinations (e.g., 350 W and 1100 mm/s). Overall, defect characteristics in CoCrMo alloy showed greater sensitivity to AM process parameters than those observed in Inconel 625. These observations not only underscore the strong influence of effective energy input on defect formation in AM materials, but also demonstrate the capability of the proposed defect detection and classification pipeline to efficiently and accurately capture such relationships. 
Its important to note that these process–defect relationships become evident only after incorporating defect type classification using etched microstructural context, as defect geometry alone is insufficient to reliably distinguish formation mechanisms. By separating defect detection from classification, the framework remains scalable and achieves good accuracy. 
Furthermore, processing-defect relations can be digested by downstream ML models to enable quick, data-driven optimization of AM process parameters for emerging alloy systems.

Despite the strong performance demonstrated by the proposed framework, several challenges remain for future work. For instance, while the SMILE-based active learning approach improves segmentation efficiency, accuracy, and reduces annotation effort, it still assumes availability of a sufficiently large and diverse unlabeled dataset. Therefore, if the initial dataset is small or highly homogeneous, the benefits of the proposed framework may reduce. Additionally, the current framework does not address scenarios where new data becomes available after model training is completed, as is common in real manufacturing environments. If newly acquired data differs significantly from the original training distribution, model performance may degrade. Thus, novel methods are needed to continuously select diverse and representative samples from evolving data distributions. Further, for the defect classification stage, the model relied on consistently and carefully prepared etched samples. Variations in etching quality, polishing or imaging noise is expected to affect the classification performance. Finally, the defect statistics produced by the proposed defect detection and classification model can be used to train surrogate ML models that map defect statistics to the AM process parameters. Such models could enable AM process optimization, even for compositions beyond the training data. Addressing these challenges will expand the applicability of the proposed framework and will allow development of a scalable and generalizable approach to predict and optimize AM process parameters.

\section{Conclusions}

Microstructure image analysis plays a fundamental role in materials science for establishing structure– property relationships. However, extracting meaningful microstructural features, such as defects, pores, and grain boundaries, requires reliable detection and classification, tasks that are often labor intensive and difficult to scale to large datasets. In this work, using an additive manufacturing (AM) dataset as a representative example, we developed a robust, efficient and scalable two-stage pipeline for defect detection and classification in microstructural images, with the primary objective of significantly reducing manual annotation effort while maintaining high model accuracy.
The first stage integrates active learning with a diversity-based sampling strategy (SMILE), U-Net based segmentation, and a human-in-the-loop framework for interactive correction/annotation. The active learning strategy enabled progressive improvement of the model as new data were incorporated, while SMILE ensured that the selected samples remained diverse and representative of the overall dataset. This strategy minimized redundant labeling and allowed the model to learn efficiently from a relatively small number of images, resulting in strong segmentation performance with macro F1 score of 0.93. From a practical standpoint, model-assisted annotation reduced labeling time by approximately 65\%, substantially lowering the overall annotation effort. 

Beyond defect detection, the framework further incorporated etched microstructural information through a dedicated second-stage classification module. By decoupling segmentation from classification, the pipeline first localized defect regions and subsequently incorporated the surrounding microstructural context to differentiate among defect types. This two-step design enabled more informed and accurate defect classification. Finally, the trained models were then used to extract quantitative defect metrics, including defect count, defect area fraction, and relative defect fractions, for the Inconel 625 and CoCrMo alloy systems. These metrics were subsequently mapped to their AM processing conditions. Importantly, the trends observed in defect statistics as a function of laser power and scan speed were found to be physically consistent with established AM process–defect relationships

Overall, the proposed pipeline provides an efficient and scalable framework for quantitative defect analysis in complex materials microstructures. While demonstrated here for AM materials, the methodology is inherently flexible and can be readily extended to other materials processing systems where reliable, scalable, and data-driven defect quantification is essential.

\section*{Supporting information}
Supporting Information is available from the Supplementary Materials. 

\section*{CRediT authorship contribution statement}
\textbf{Sanjeev S. Navaratna:} Data curation, Methodology, Software, Conceptualization, Formal analysis, Writing -- original draft.  \textbf{Nikhil Thawari:} Data curation, Investigation, Writing -- review \& editing. \textbf{Gunashekhar Mari:} Conceptualization, Data curation, Writing -- review \& editing. \textbf{Amritha V P:} Data curation, Writing -- review \& editing. \textbf{Murugaiyan Amirthalingam:} Conceptualization, Resources, Supervision, Writing -- review \& editing. \textbf{Rohit Batra:} Conceptualization, Resources, Supervision, Project administration, Writing -- review \& editing, Funding acquisition.

\section*{Declaration of Competing Interest}
The authors declare no competing interests.

\section*{Acknowledgments}
Rohit Batra acknowledges ANRF (File No. SRG/2023/001055), DRDO-DIA and WSAI for funding. Computational
resources from Robert Bosch Centre for Data Science and AI, IIT Madras, are also acknowledged. Sanjeev S. Navaratna gratefully acknowledges Vishaal Harikrishna Kumar and Sumit Katiyar for their assistance in the preparation of the etchant solution.

\section*{Data Availability}
The data and code that support the findings of this study are publicly available at this \href{https://github.com/MI-LAB-IITM/Semi_Automated_AM_Defect_Analysis_Framework}{Github repository}. The repository contains the different datasets, core-set SMILE algorithm implementation, model training and testing codes, and CVAT-based annotation implementation code used for defect detection and defect classification pipeline.

\bibliography{ref}

\end{document}